# Single photon in hierarchical architecture for physical reinforcement learning: Photon intelligence

Makoto Naruse[1,*], Martin Berthel[2,3], Aurélien Drezet[2,3], Serge Huant[2,3],

Hirokazu Hori[4] and Song-Ju Kim[5]

[1] Network System Research Institute, National Institute of Information and Communications Technology, 4-2-1 Nukui-kita, Koganei, Tokyo 184-8795, Japan

[2] CNRS, Inst. NEEL, F-38042 Grenoble, France

[3] Université Grenoble Alpes, Inst. NEEL, F-38000 Grenoble, France

[4] Interdisciplinary Graduate School of Medicine and Engineering, University of Yamanashi, Takeda, Kofu, Yamanashi 400-8511, Japan

[5] WPI Center for Materials Nanoarchitectonics, National Institute for Materials Science, 1-1 Namiki, Tsukuba, Ibaraki 305-0044, Japan

* Corresponding author at: 4-2-1 Nukui-kita, Koganei, Tokyo 184-8795, Japan.

*Email address:* naruse@nict.go.jp

**ABSTRACT**

Understanding and using natural processes for intelligent functionalities, referred to as natural intelligence, has recently attracted interest from a variety of fields, including post-silicon computing for artificial intelligence and decision making in the behavioural sciences. In a past study, we successfully used the wave-particle duality of single photons to solve the two-armed bandit problem, which constitutes the foundation of reinforcement learning and decision making. In this study, we propose and confirm a hierarchical architecture for single-photon-based reinforcement learning and decision making that verifies the scalability of the principle. Specifically, the four-armed bandit problem is solved given zero prior knowledge in a two-layer hierarchical architecture, where polarization is autonomously adapted in order to effect adequate decision making using single-photon measurements. In the hierarchical structure, the notion of layer-dependent decisions emerges. The optimal solutions in the coarse layer and in the fine layer, however, conflict with each other in some contradictive problems. We show that while what we call a *tournament* strategy resolves such contradictions, the probabilistic nature of single photons allows for the *direct location of* the optimal solution even for contradictive problems, hence manifesting the exploration ability of single photons. This study provides insights into photon intelligence in hierarchical architectures for future artificial intelligence as well as the potential of natural processes for intelligent functionalities.

Modern society is becoming increasingly reliant on artificial intelligence (AI)[1]. AI at present is based on computer algorithms and digital computing, and suffers from a theoretical limitation known as the von Neumann bottleneck[2,3], as the design of conventional digital computing devices anticipates the end of Moore's law[4], which imposes limits on the extent to which integrated circuits can be downscaled. Consequently, the utilization of unconventional physical processes and architectures for intelligence, referred to as natural intelligence, is attracting increasing attention. The relevant methods include quantum annealing[5], laser-based solution search[6], new type of solution-searching circuits such as complementary metal–oxide–semiconductor (CMOS) annealing[7] and the photon intelligence approaches proposed by us[8,9]. Meanwhile, human intelligence, especially decision making, has been intensively examined through such mathematical and physical modelling approaches as quantum decision theories[10,11] and neuroscience[12], part of which influences reinforcement learning algorithms[13]. Physical insights into and the implementation of natural intelligence, particularly with regard to decision making, are stimulating for computation, physics, and the behavioural sciences.

One of the most fundamental issues in machine learning and the decision sciences is the multi-armed bandit problem (MAB), which concerns how to maximize the total reward from multiple slot machines[14]. To solve this problem in general, an exploration of search procedures for the highest-reward probability machine, which is precisely defined shortly below, is needed; however, too much exploration may result in excessive loss, whereas too quick a decision or insufficient exploration might result in missing the best machine. This is called the "exploration-exploitation dilemma"[14–16]. The MAB is important for various practical applications, such as information network management[17,18], Web advertisement[19], Monte Carlo tree search[20], and clinical trials[21].

In our previous study, we experimentally showed that a single photon can solve the two-armed bandit problem using the nitrogen-vacancy (NV) centre in a nanodiamond as a single-photon source[9]. The wave-particle duality of the single photon is utilized where the probabilistic attribute of the photon takes the role of exploration while its particle nature is immediately and directly associated with a particular decision. The theoretical background for this has been examined by comparisons to other reinforcement learning algorithms[22] and category-theoretic modelling and analysis[23].

However, many issues remain unresolved in the route to realizing artificially constructed, physical decision-making machines. A fundamental issue in this regard is scalability; the number of choices involved in a decision may be numerous, not merely binary as assumed in the first proof-of-principle experiment in Ref. 9. A hierarchical architecture is a promising approach to scalable intelligent systems[24] and optical devices dealing with multiple channels[25], and has been applied to quantum computing platforms[26].

The effectiveness of a hierarchical approach for single-photon-based decision making is, however, completely unknown. Moreover, interesting notions of decision making emerge in a hierarchical architecture—that is, decision making at *finer* and *coarser* scales of the hierarchy. Specifically, this paper shows that the four-armed bandit problem is resolved given zero prior knowledge by using single photons in a two-layer tree-structure architecture, where the polarization of single photons is autonomously adapted. We have to be aware that the optimal solution in the coarse layer can conflict with that in the fine layer (as explained in detail below in the problem exemplified by `CASE 3`). We show that while a simple "tournament" strategy resolves such contradictions, the probabilistic nature of single photon allows the *direct location* of the optimal solution on a fine scale, which is a manifestation of the exploration ability of single photons. This manifests yet another fundamental of the quantum nature of single photons in vital intelligent roles, in contrast to the literature on single photons that focuses only on the contexts of quantum key distributions[27] and quantum computing[28,29].

**Results**

**Hierarchical single-photon-based decision maker.** For the simplest case that preserves the essence of the solution of the MAB problem in a hierarchical system, we consider a player who selects one of four slot machines (**slot machines 1**, **2**, **3** and **4**) with the goal of maximizing a reward. Denoting the reward probabilities of the slot machines by $P_i$ $(i = 1, \cdots, 4)$, respectively, the problem then is to select the machine with the highest reward probability, referred to as the 'highest-reward probability machine'. A unit reward dispensed by a slot machine is identical among all machines. The machine-selection decision is associated with single-photon detection by designated photo detectors corresponding, respectively, to **slot machines 1** to **4**, as described below.

The architecture of the optical system has a tree structure, where an incoming single photon is directed by a polarizing beam splitter (PBS), denoted by PBS$_1$ in Fig. 1, following which it experiences another PBS, either PBS$_2$ or PBS$_3$, resulting in photon detection by one of four avalanche photodiodes (APDs), APD$i$ ($i = 1, \cdots, 4$) in Fig. 1. The central idea of single-photon decision making is to adapt the polarization of single photons by the angle of three half-wave plates (denoted by HWP$_1$, HWP$_2$, HWP$_3$) located at the fronts of three PBSs, respectively.

We see here a hierarchical structure: PBS$_1$ governs the decision of whether to select {**slot machine 1** or **2**} or {**slot machine 3** or **4**}, referred to as the "coarse-scale" decision hereafter. For the sake of simplicity, we call the relevant collections *Group 1* for {**slot machine 1** and **2**} and *Group 2* for {**slot machine 3** and **4**}. That is, the coarse-scale decision concerns whether to choose either of **Group 1** or **Group 2**: namely, which of the reward probabilities $P_1 + P_2$ or $P_3 + P_4$ has the greater value? Meanwhile, PBS$_2$ and PBS$_3$ concern the machine selection decision concerning [**slot machine 1** or **2**] and [**slot machine 3** or **4**], respectively. We call a "fine scale" decision one that concerns the choice of a machine with the maximum $P_i$ ($i = 1, \cdots, 4$).

Initially, the linear polarization of the single photons that are input is oriented at $\pi/4$ with respect to the horizontal of PBS$_1$, enabling the photons to be directed to **Group 1** or **Group 2** with 50:50 probability. We suppose that due to PBS$_1$, the polarization of the photon directed towards **Group 1** is *vertically* polarized, whereas that directed towards **Group 2** is *horizontally* polarized. The polarizations of single-photon incidents on PBS$_2$ and PBS$_3$ are also initially orientated at $\pi/4$ with respect to the horizontal of PBS$_2$ and PBS$_3$ via HWP$_2$ and HWP$_3$, respectively; hence, photons are to be directed towards APD$_1$ or APD$_2$ with a 50:50 probability whereas those directed towards APD$_3$ or APD$_4$ have the same probability. Meanwhile, the *total* probability of photon detection by either of the APD$_i$ ($i=1,\cdots,4$) is 1. This is a notable aspect of the single-photon decision maker in the sense that the probabilistic (wave) and particle attributes of a single photon are employed.

The principle of single-photon-based decision making is inspired by the tug-of-war (TOW) method invented by Kim *et al.*[22,30], which originated from the observation of slime moulds—the concurrent expanding and shrinking of their bodies, while maintaining a constant intracellular resource volume, allows them to gather environmental information, and the conservation of the volumes of their bodies entails a nonlocal correlation among the body. The TOW is a metaphor to represent such nonlocal correlation, which enhances decision making performance[22].

This mechanism satisfactorily matches the intrinsic attributes of a single photon in a hierarchical architecture, as examined in this study, not merely a two-armed system[9]. Until a single photon is detected by either of multiple detectors, a single photon is *not* localized in the system. The possibility of photon detection at each of the APDs is *not* perfectly zero unless single photon polarization is *perfectly* horizontal or vertical. This is a remarkable aspect of the single-photon decision maker: it exploits the *quantum* attributes of photons. If photon observation was based on *classical* light, e.g. observing light intensity, we would have needed to implement an additional step to facilitate decision making.

In our hierarchical single-photon-based decision making architecture, the TOW mechanism is implemented by three *polarization adjusters* (PAs), which are respectively marked PA$_i$ ($i = 1, 2, 3$), and control the corresponding HWP$_i$ ($i = 1, 2, 3$). The numerical indicators of the polarization adjusters, referred to as the *PA value* hereafter, are also represented by $PA_i$ ($i = 1, 2, 3$). The control mechanisms of the PAs for the coarse and fine scales are given by the following:

**Control Mechanism of PA$_1$ (coarse-scale control)**

[C-1] A PA$_1$ value of *zero* indicates a polarization at 45° with respect to the horizontal; single photons are directed towards **Group 1** or **2** with 50:50 probability.

[C-2] The decision to select the slot machine is immediately made by observing a single photon in APD$i$ ($i = 1, \cdots, 4$).

[C-3] If a reward is successfully dispensed from one of the slot machines in **Group 1**, PA$_1$ is "moved" in the direction of the chosen **Group**, i.e. if **slot machine 1 or 2** is selected based on photon detection by APD$_1$ or APD$_2$ and a reward is obtained, PA$_1$ is moved such that input polarization is more *horizontally* polarized by controlling HWP$_1$. Moreover, if no reward is dispensed by **slot machine 1 or 2**, the PA is moved in the direction of the unselected machine, i.e. input polarization is more *vertically* polarized in this case. The same mechanism applies to **Group 2** with **slot machines 3** and **4**.

By iterating steps **[C-2]** and **[C-3]**, PA$_1$ configures the system so that the **group** with the higher probability of reward is more likely to be selected by incoming single photons. The details of the mechanism are formulated below (**TOW mechanism of PA$_1$**).

**Control Mechanism of PA$_2$ and PA$_3$ (fine-scale control)**

**[F-1]** A PA$_2$ value of zero indicates a polarization at 45° with respect to the horizontal; single photons, impinging on PBS$_2$, are directed to APD$_1$ or APD$_2$ with 50% probability.

**[F-2]** The decision to select the slot machine is immediately made by observing a single photon in APD$i$ $(i = 1, \cdots, 4)$.

**[F-3]** If a reward is successfully dispensed by **slot machine 1**, PA$_2$ is moved in the direction of **slot machine 1**, i.e. if **slot machine 1** is selected and a reward is obtained, PA$_2$ is moved such that input polarization is more *horizontally* polarized by controlling HWP$_2$. Moreover, if no reward is dispensed by **slot machine 1**, PA$_2$ is moved in the direction of the unselected machine, i.e. input polarization is more *vertically* polarized. The same mechanism applies to **slot machine 2**.

By iterating steps **[F-2]** and **[F-3]**, PA$_2$ configures the system such that the machine with the higher reward probability between **slot machines 1** and **2** is more likely be selected. The same architecture is adapted for PA$_3$ with respect to **slot machines 3** and **4**. The details of the mechanism are formulated below (**TOW mechanism of PA$_2$ and PA$_3$ (fine-scale TOW)**).

**Measurements.** The experimental setup was based on the architecture shown in Fig. 1. A single photon was emitted by a nitrogen-vacancy (NV) colour centre from a nanodiamond[31], which featured broadband emission in the visible range (650-700 nm) at room temperature[32], passed through a polarizer and a zero-order half-wave plate (HWP$_1$), and impinged on PBS$_1$. One of the branches connected to another zero-order half-wave plate (HWP$_2$) followed by PBS$_2$, whereas the other branch connected to HWP$_3$ and PBS$_3$. The orientations of HWP$_i$ $(i=1,\cdots,3)$ were configured by respective rotary servomotors. The single photon was detected by one of four APDs (APD$i$ $(i=1,\cdots,4)$) connected to a 100-ps bin size and a multiple-event time digitizer (time-to-digital converter (TDC)) to record detection times. The details of the optical system used in the experiment and the single-photon emission from the NV centre are described in the *Methods* section and in the Supplementary Figures.

Example sequences of single-photon series detected by APDs are shown in Figs. 2a, 2b and 2c. The vertical red, green, blue and cyan bars represent single-photon detection events by APD$i$ $(i=1,\cdots,4)$, and the horizontal axis represents time. In Fig. 2a, all HWPs are configured so that the output polarizations are approximately 45° with respect to the horizontal and the incoming single photons are directed to the four APDs with equal probability. On the contrary, in Fig. 2b, the HWP$_1$ configures the polarization nearly perfectly horizontally; hence, detection is induced mostly by either APD$_1$ or APD$_2$. In Fig. 2c, HWP$_2$ is also configured so that the output polarization is nearly horizontal, leading to detection mostly by APD$_1$. Note, however, that a few photons are observed in, for example, APD$_4$ in Fig. 2b and APD$_2$ in Fig. 2c; such *probabilistically rare* events are important for adaptive decision making in uncertain environments[22].

In order to play a slot machine and reconfigure HWPs, some time is needed. In this study, the *decision* is made by the *first single-photon detection* event of a given cycle; if detection occurs at APD$_i$, the decision is *immediately* made to play the **slot machine i**. In the experiment, the slot machines were emulated by a host controller (See the *Methods* section for details). Specifically, reward probabilities $P_i$ $(i=1,\cdots,4)$ were given as threshold values. If a random number between 0 and 1 generated by the host controller was less than the reward probability of the selected slot machine, reward was dispensed.

Based on the PA values ($PA_i$ $(i=1,\cdots,3)$), the linear polarization was made more vertical or horizontal by rotating the HWPs using a rotary positioner. In Figs. 2d, 2e and 2f, the red, green, blue and cyan circles, respectively, indicated 0.5 s worth of photon counts detected by APD$_i$, $(i=1,\cdots,4)$, as a function of the orientation of the half-wave plates. In controlling HWP$_i$, other HWPs (HWP$_j$ $(j \neq i)$) were kept in 50:50 setups. Note that the orientation angles shown in the horizontal axis did not indicate linear polarization with respect to the horizontal direction, but to the absolute value defined in the rotary positioner used in the experiment. We clearly observed that (1) the single photon incidence followed similar characteristics in **Group 1** (APD$_1$ and APD$_2$) and **Group 2** (APD$_3$ and APD$_4$) with respects to HWP$_1$ dependencies (Fig. 2d), (2) whereas [APD$_1$ and APD$_2$] and [APD$_3$ and APD$_4$] exhibited an opposite trend regarding HWP$_2$ and HWP$_3$, respectively (Fig. 2e and Fig. 2f).

Because the sensitivities of the APDs were not identical, and owing to possible misalignment in the optical setup, the polarization dependencies did not exhibit perfect symmetry. The extinction ratio of the polarizer was $10^5$ and that of the PBS was $10^3$ (product information is shown in the *Methods* section). We think that the intrinsic optical properties of various optical components in the experimental setup did not yield significant asymmetry.

To implement the PA mechanisms in the hierarchical architecture, we quantified the TOW mechanism as below. Let all initial PA values be zero.

**TOW mechanism of PA$_1$ (coarse-scale TOW)**

If, in cycle *t*, the selected machine yields a reward (or in other words, the slot machine wins), PA$_1$ is updated at cycle $t+1$ based on

$$\begin{aligned} PA_1(t+1) &= -\Delta_1 + \alpha_1 PA_1(t) \quad \text{if \textbf{slot machine 1} or \textbf{2} (\textbf{Group 1}) wins} \\ PA_1(t+1) &= +\Delta_1 + \alpha_1 PA_1(t) \quad \text{if \textbf{slot machine 3} or \textbf{4} (\textbf{Group 2}) wins} \end{aligned} \quad (1)$$

where $\alpha_1$ refers to the forgetting parameter[8], and $\Delta$ is the constant increment (in this experiment, $\Delta_1 = 1$ and $\alpha_1 = 0.999$). When the selected machine does *not* yield a reward (or loses in the slot play), PA$_1$ is updated by

$$\begin{aligned} PA_1(t+1) &= +\Omega_1 + \alpha_1 PA_1(t) \quad \text{if \textbf{slot machine 1} or \textbf{2} (\textbf{Group 1}) loses} \\ PA_1(t+1) &= -\Omega_1 + \alpha_1 PA_1(t) \quad \text{if \textbf{slot machine 3} or \textbf{4} (\textbf{Group 2}) loses} \end{aligned} \quad (2)$$

where $\Omega_1$ is a parameter defined below. Intuitively speaking, *PA$_1$ increases* if the slot machines in **Group 1** are more likely to win, and *decreases* if those in **Group 2** are considered to be more likely to earn rewards. This is as if the value of $PA_1$ is being pulled by **Group 1** and **Group 2**, which coincides with the notion of TOW. The values of $PA_1$ is then adapted to polarization control via HWP1, so that polarization is more horizontal in the former case (**Group 1** is more likely to win) and vertical in the latter (**Group 2** is more likely to win). Specifically, the orientation of HWP1 at cycle $t$ is determined by

$$HWP_1(t) = POS_1\left(\lceil PA_1(t) \rceil\right) \tag{3}$$

where $\lceil\ \rceil$ represents the round-off function to the closest whole number. Function $POS_1(n)$ specifies the orientation of HWP1 based on polarization dependencies characterized as in Fig. 2d (details are described in the *Methods* section).

In TOW-based decision making, $\Omega_1$ is determined based on the history of betting results. Let *the number of* **slot machines selected** *i* by cycle $t$ be $N_i$ and *the number of winning* **slot machines** *i* be $L_i$. The estimated reward probabilities by the slot machines in **Group 1** and **Group 2** are, respectively, given by

$$\hat{P}_{G1} = \frac{L_1 + L_2}{N_1 + N_2},\ \hat{P}_{G2} = \frac{L_3 + L_4}{N_3 + N_4}. \tag{4}$$

$\Omega_1$ is then given by

$$\Omega_1 = \frac{\hat{P}_{G1} + \hat{P}_{G2}}{2 - (\hat{P}_{G1} + \hat{P}_{G2})} \tag{5}$$

while the initial $\Omega_1$ value is assumed to be unity, and a constant value is assumed when the denominator of Eq. (5) is zero. The detailed derivation of Eq. (5) is shown in Ref. 22.

**TOW mechanism of PA$_2$ and PA$_3$ (fine-scale TOW)**

We describe only the TOW mechanism of PA$_2$ below, since that of PA$_3$ follows the same principle and the corresponding slot machines. If, in cycle $t$, the selected machine yields a reward (or in other words, wins the bet), the value of PA$_2$ is updated at cycle $t+1$ based on

$$PA_2(t+1) = -\Delta_2 + \alpha_2 PA_2(t) \quad \text{if \textbf{slot machine 1} wins}$$
$$PA_2(t+1) = +\Delta_2 + \alpha_2 PA_2(t) \quad \text{if \textbf{slot machine 2} wins} \tag{6}$$

with $\Delta_2 = 1$ and $\alpha_2 = 0.999$, as in the case of $PA_1(t)$. When the selected machine does *not* yield a reward, the value of PA is updated by

$$PA_2(t+1) = +\Omega_2 + \alpha_2 PA_2(t) \quad \text{if \textbf{slot machine 1} loses}$$
$$PA_2(t+1) = -\Omega_2 + \alpha_2 PA_2(t) \quad \text{if \textbf{slot machine 2} loses} \tag{7}$$

where $\Omega_2$ is a parameter defined later (Eq. (10)). In other words, the value of PA$_2$ is pulled by **slot machines 1** and **2** in a tug-of-war manner. The value of PA$_2$ is adapted to HWP$_2$ control so that polarization is more horizontal if **slot machine 1** is expected to dispense more reward, whereas it is more vertical if **slot machine 2** is considered to be more beneficial. As in the former case, the orientation of HWP$_2$ is determined by

$$HWP_2(t) = \text{POS}_2\left(\lceil PA_2(t) \rceil\right) \tag{8}$$

where $POS_2(n)$ specifies the orientation of HWP₂ based on the polarization dependencies characterized in Fig. 2e.

The estimated reward probabilities of **slot machine 1** and **slot machine 2** are, respectively, given by

$$\hat{P}_1 = \frac{L_1}{N_1}, \quad \hat{P}_2 = \frac{L_2}{N_2}, \tag{9}$$

followed by $\Omega_2$, which is given by

$$\Omega_2 = \frac{\hat{P}_1 + \hat{P}_2}{2 - (\hat{P}_1 + \hat{P}_2)}. \tag{10}$$

As mentioned earlier, the other PA value in the fine scale, $PA_3(t)$, is given in the same manner, by taking account of **slot machines 3** and **4** instead of **slot machines 1** and **2**.

The decision-making procedure in the hierarchical architecture is summarized as follows:

[1] Photon arrival time is measured through APDs and a TDC system.

[2] The decision is made based on the first photon detection in Step [1]. The selected **slot machine** is played.

[3] Reward is dispensed or not.

[4] The PA values are updated based on Eqs. (1), (2), (6) and (7).

[5] The orientations of the HWPs are determined using Eqs. (3) and (8).

[6] The values of $\Omega_i$ $(i = 1, \cdots, 3)$ are updated using Eqs. (4), (5), (9) and (10).

[7] The rotary positioner is controlled; then, the system returns to Step [1].

**Discussion**

We first solve the typical four-armed bandit problems given by following two cases, where the reward probabilities are respectively given by

$$\begin{aligned}\text{CASE 1}: \{P_1, P_2, P_3, P_4\} &= \{0.8, 0.2, 0.1, 0.1\} \\ \text{CASE 2}: \{P_1, P_2, P_3, P_4\} &= \{0.8, 0.1, 0.2, 0.1\}\end{aligned}. \qquad (11)$$

Since the maximum reward probability is $P_1 = 0.8$ for both cases, the correct decision (precisely speaking, the correct decision in the fine scale) is to select **slot machine 1**. Note also that the elements of the probabilities are identical, although the order is slightly different.

Starting with *zero* prior knowledge of the reward probabilities, the hierarchical single-photon-based decision maker makes consecutive 30 plays and repeats these plays 10 times. The red solid and dashed lines in Fig. 3a, respectively, show the correct decision rate at cycle *t* for `CASE 1` and `CASE 2` problems, defined by the ratio of the number of time the highest-reward probability machine is chosen in cycle *t* in all trials (10 times), which gradually increases over time, demonstrating successful decision making.

Since $P_1 + P_2 > P_3 + P_4$ holds for both cases, choosing machines in **Group 1** is the correct decision at the coarse scale. We note that the *difference* of the sum of reward probabilities between **Group 1** and **Group 2** is given by `[CASE1]` $(P_1 + P_2) - (P_3 + P_4) = 0.8$ and `[CASE2]` 0.6, indicating that `CASE 1` is a relatively easier problem than `CASE 2`. Indeed, the blue solid and dashed lines in Fig. 3a show the correct decision rate at the coarse scale for `Cases 1 and 2`, respectively, defined by the ratio of the number of selections of machines belonging to **Group 1** in all trials, where `Case 1` is quickly approaching unity, namely, the case where rapid adaptation is implemented. Consequently, `Case 1` exhibits more rapid adaptation than `Case 2` at the fine scale as well, as shown by the red lines in Fig. 3a. Figure 3b summarizes the temporal evolution of polarization adjuster values. Here, we observe that

(1) Both $PA_1$ and $PA_2$ in `Case 1` (red and blue solid curves, respectively) decrease more quickly than those in `Case 2` (red and blue dashed curves, respectively), meaning that single-photon polarization is shifted to the horizontal by HWP$_1$ and HWP$_2$, coinciding with the decision-making performance in Fig. 3a.

(2) $PA_3$, which concerns decisions among **slot machines 3** and **4**, persists with a value of around zero in both cases (green solid and dashed curves for `Case 1` and `Case 2`, respectively), since the difference in their reward probabilities is zero or very small; hence, these machines are rarely selected when the system finds the best machine.

In these demonstrations as well as the following ones, the resolutions of polarization control specified by $HWP_i(t)$ $(i=1,\cdots,3)$ consist of seven steps.

The highest-reward probability slot machine may *not* belong to the higher-reward probability group at the coarse scale. Take, for example, a case given by

$$[\text{Case 3}]\ \{P_1, P_2, P_3, P_4\} = \{0.7, 0.5, 0.9, 0.1\} \tag{12}$$

where the correct decision is to choose **slot machine 3** ($P_3 = 0.9$), which belongs to **Group 2**. At the coarse scale, on the contrary, **Group 1** has a larger winning probability ($P_1 + P_2 = 1.2$) than **Group 2** ($P_3 + P_4 = 1.0$), which implies that the optimal solution at the coarse scale is to select either **slot machine 1** or **2**, which is *contradictory* to the correct decision at the fine scale (**slot machine 3**). Hence, polarization control at the coarse scale (HWP₁) encounters difficulty in guiding the decision towards **Group 2**. The red solid line in Fig. 4 shows the correct selection rate at the fine scale, which fluctuates about 0.5, indicating that the system was *not* able to find the correct decision.

One way to derive the correct decision at the fine scale for such a contradiction problem is what is called the "tournament" method: In the *first round*, coarse-scale polarization control is *not* activated ($PA_1$ is kept zero) while each branch of the fine scale experiences polarization adjustments. As a consequence, a higher-reward probability machine is more likely to be chosen at each branch, that is, $P_1 = 0.7$ in **Group 1** and $P_3 = 0.9$ in **Group 2**. In the *second round*, coarse-scale polarization control is initiated, whereas fine-scale control is fixed; this means that the TOW principle applies to the *winners of the first round*, leading to the highest-probability machine at the end. The blue lines in Fig. 4 show experimental verification based on the tournament method, where the first 15 cycles belonged to the first round and the next 15 to the second round. In the second round, we observed that the correct decision rate at the finer scale, depicted by the solid blue line, increased, whereas that at the coarse scale, shown by the dashed blue line, decreased, unlike in the "non-tournament" method discussed earlier, and represented by red lines; this showed that the tournament method works successfully in a single-photon-based hierarchical decision maker.

The other approach to derive the global optimal solution, or the maximum-reward probability machine, in the hierarchical architecture *without* using tournament-based approaches involves *further exploiting* the probabilistic attributes of photons by increasing the resolutions of the polarization adjuster. Concerning the same problem `[Case 3]`, the solid, dashed, dotted and dash-dotted lines in Fig. 5a demonstrate correct decision rates (at the fine scale) with the *number of resolutions of polarizations*, or the *number of steps in polarization adjustment*, being 11, 9, 7 and 5, respectively. We can observe that with increasing resolution, the correct decision rate at the fine layer increases. This is also shown by the square marks in Fig. 3b that compare the correct decision rates at cycle 30 as a function of the number of polarization adjustment steps. We should also remark that the correct decision rate at the coarse layer decreases with increasing number of PA steps, as shown by the square marks in Fig. 3b. The number of steps of polarization adjustment resolution can have a much larger value; Figure 5c shows numerical simulations of the correct decision rate until cycle $t = 500$, calculated as an average of 100 iterations with varying numbers of polarization adjustment steps: 5, 7, 9, 11, 51 and 101. With increasing resolution, the correct decision rate approached unity, whereas the adaptation became slower. At cycle $t = 30$, the correct decision rate at the fine and the coarse scales behaved as the number of polarization adjustment steps, as shown in Fig. 4d; the portion between resolutions 5 and 11 exhibited a similar trend in the experimental results shown in Fig. 5b, demonstrating the exploration abilities of single photons for the optimal solution in a hierarchical architecture.

Finally, we discuss this study in light of practical considerations. First, the second-order photon-intensity correlation $g^{(2)}(0)$ of the single-photon source was sufficiently smaller than unity, as shown in Supplementary Fig. 2. Therefore, we observed *no* events where photons were detected by multiple APDs in the same time bin during the decision-making experiments. If multiple APDs simultaneously detect photons, a decision *cannot* be made, and the system needs to try another photon observation. Meanwhile, since the single-photon generation rate in our experiment from the NV in a nanodiamond was approximately 50 k photons/s, there was *no* photon observation in *any* channel in experiments where a single capture of 100-ps timing resolution detection spanned only approximately 10 μs due to limited bandwidth between the TDC and the host controller; hence, we repeated photon measurements until a single photon was observed. This is among the practical limits of the operating speed of the system. The host controller (see the *Methods* section for specifications) serially controls three rotary servomotors for HWPs. Overall, the latency for a single slot play spanned around a few seconds, which limited the number of executable slot plays in the experiments. Improving the operating speed of the total system by enhancing the single-photon rate[33], the speed of the polarization modulation by, for example, electro-optic phase modulators and incorporating fully parallel control mechanisms are important engineering topics for future research.

Second, integration is an important issue. Adapting integrated planar lightwave circuit technologies, previously used in some quantum systems[26,34], into photon intelligence is an interesting topic for future study. Another possibility is the use of nanophotonic devices based on optical near-field–mediated energy transfer[35] based on quantum dots[36] and shape-engineered nanostructures[37].

Third, there is the need to study more sophisticated decision-making problems. This study dealt with the multi-armed bandit problem for a *single* player. As clearly shown in Fig. 4c, there is a trade-off between the speed of adaptation and the accuracy of decisions; examining an *optimal strategy* based on the requirements of applications is an interesting topic of research. Moreover, when the number of players increases to more than one, called the competitive multi-armed bandit (CMAB) problem, the entire problem becomes more complicated and the issue of the Nash equilibrium becomes a serious concern[38]. Kim *et al.* proposed a physical architecture that solves CMABs[39]. Using the intrinsic attributes of photons, such as entanglement[28], for such complex problems is another exciting research topic in photon intelligence.

To summarize, we have experimentally demonstrated that single photons in hierarchical architectures can solve multi-armed bandit problems from zero prior knowledge for reinforcement learning and decision making based on the intrinsic wave-particle duality of photons, which is a decisive step towards verifying an important architecture for massive parallelism. Using the NV centre in a nanodiamond as a single-photon source, the polarization of single photons was adapted by multiple polarization adjustors such that the highest-reward probability machine was chosen. The notions of coarse- and fine-scale decisions emerge in hierarchical architectures. The correct decision at the coarse and fine scales may contradict with each other. We showed that while the global optimal solution was derived by solving from the finer to the coarser scale in a step-by-step manner, referred to as the tournament method, exploiting the probabilistic abilities of a single photon allows the direct derivation of the optimal selection. This study unveiled single-photon intelligence in a hierarchical architecture for future AI as well as the potential of natural processes for functionalities.

**Methods**

**Optical system.** The single-photon beam was supplied by a single NV centre in a surface-purified 80-nm nanodiamond[40]. The image shown in Supplementary Fig. 1 is a confocal microscope image of the single NV centre, and the blue curve in Supplementary Fig. 2 shows the result of measuring second-order photon correlation from the single NV centre. The measurement was made using a standard Hanbury–Brown and Twiss correlator. The red curve is a fit with a three-level model[41]. The antibunching dip, which is a signature of the quantum nature of the emission, did drop to about 0.2 with a finite delay. The excitation laser was operated at a wavelength of 532 nm (green diode-pumped solid-state laser by CNI Lasers). The four detectors consisted of APDs (Excelitas Technologies SPCM-AQRH-16-FC), with a peak photon detection efficiency greater than 70% at 700 nm and a background count below 25 counts per second. The TDC system was a FAST ComTec MCS6A with a minimum time bin width of 100 ps. The three half-wave plates were mounted on Thorlabs motorized rotary positioners (PRM1Z8) driven via DC servomotors at a resolution of 1 arc-sec. We used standard optical components adapted to our spectral range [three PBSs (CM1-PBS252), three half-wave plates (WPH10M-694), and a polarizer (LPVISB050-MP2) from Thorlabs, a long-pass filter (BLP02-561R-25; cutoff wavelength: 561 nm) from Semrock and an immersion objective (numerical aperture: 1.4) from Nikon (CFI Pal Apo VC 100X Oil)]. The host controller was Hewlett Packard's HP Z400 with an Intel Xeon W3520 CPU (2.66 GHz), OS Windows 7 professional 64 bits and 4 GB of RAM. A LabVIEW (version 2015) program was used to control the experimental system, including the emulation of slot machines.

**Polarization Adjusters.** For decision-making experiments for all cases (Cases 1, 2, 3), the rounded integer values of the PA values, $\lceil PA_i(t) \rceil$ ($i=1,2,3$) shown in Eqs. (3) and (8), were assumed to be elements of the set −3, −2, −1, 0, 1, 2, 3. That is, there were seven polarization adaptation steps. When $PA_i(t) \geq 4$ or $PA_i(t) \leq -4$, the rounded PA value was defined by $\lceil PA_i(t) \rceil = -3$ and $\lceil PA_i(t) \rceil = 3$, respectively. Note that the PA value itself, $PA_i(t)$, can take any real number value.

Let the orientation of the linear polarization based on the rounded integer values of zero be given by $POS_i(0) = 0$; the other PA values are assigned by, with a constant angle interval $45°/(7-1) = 7.5°$, $POS_i(n) = -7.5 \times n$ ($n=-3,\cdots,3$). In the experimental and numerical evaluations of the decision making in Case3, other intervals of polarization were employed by $S = 45°/(N-1)$, where $N$ is an odd number representing the number of polarization adaptation steps, leading to $POS_i(n) = -S \times n$ ($n = -\lfloor N/2 \rfloor, \cdots, \lfloor N/2 \rfloor$).

**Acknowledgements**

The nanodiamond sample was provided by G. Dantelle and T. Gacoin. This work was supported in part by the Core-to-Core Program, A. Advanced Research Networks from the Japan Society for the Promotion of Science, and in part by Agence Nationale de la Recherche, France, through the SINPHONIE project (Grant No. ANR-12-NANO-0019).

**Author contributions**

M.N., S.H. and S.-J.K. directed the project. M.N. and S.-J.K. designed the system architecture. M.N., M.B., A.D. and S.H designed and implemented the optical systems. M.B. and M.N. conducted the optical experiments. M.N. and S.-J.K. analysed the data. H.H. discussed the physical principles, and M.N., M.B., A.D. and S.H. wrote the paper.

**Competing financial interests**

The authors declare no competing financial interests.

**Author Information**

Correspondence should be addressed to M.N. (naruse@nict.go.jp).

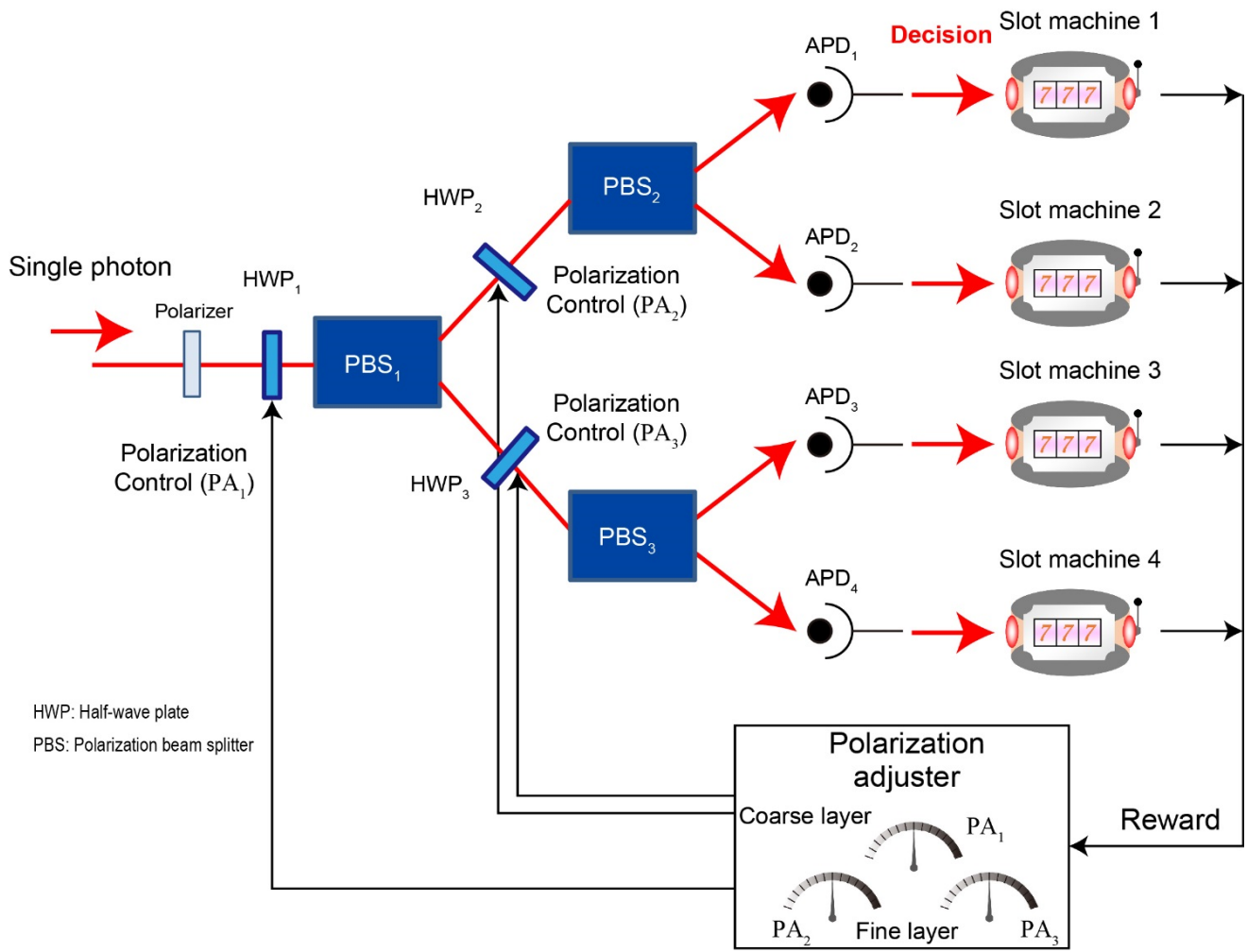

**Figure 1 | Single photon in hierarchical architecture for physical reinforcement learning.** Linearly polarized single photons, emitted from the NV centre in a nanodiamond, are directed to one of four photodetectors through three half-wave plates and polarized beam splitters arranged in a hierarchical or tree structure. The detection event at each detector is immediately associated with the selection of the slot machine. Due to the wave attribute of the single photon, the probability of photon detection for any of the detectors may not be zero, whereas individual single photon results in detection at one of the detectors thanks to the particle nature of photons. On the basis of the betting results, three polarization adjusters (PAs) controlled the orientation of the half-wave plates using rotary positioners, so that the single photon was more likely to be detected by the higher-reward-probability slot machine.

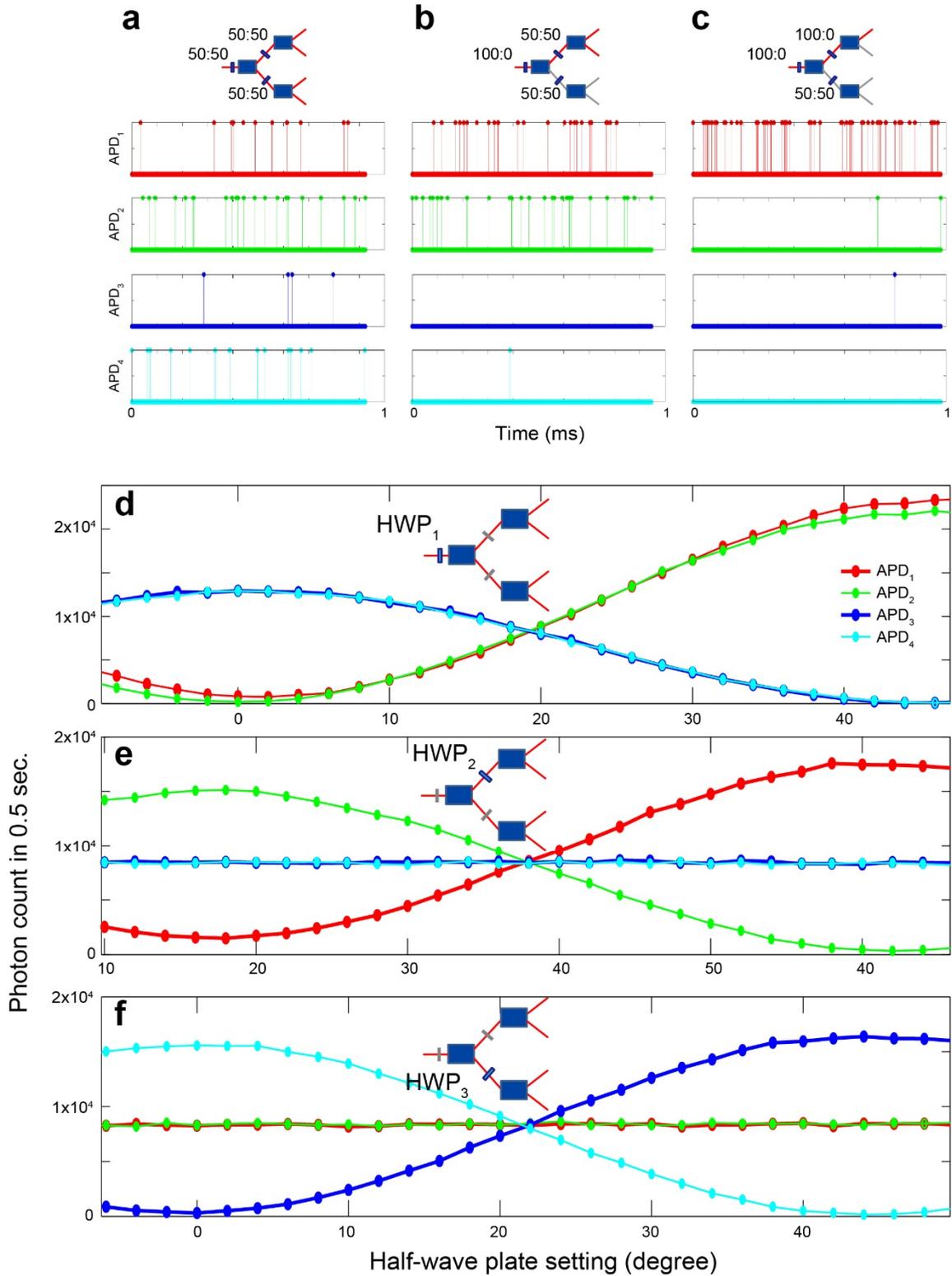

**Figure 2 | Series of single photons in the hierarchical architecture depending on the polarizations. (a, b, c)** Single-photon series over about 1 ms detected by ADSs with different HWP setups. **(d, e, f)** Photon counts over 0.5 s as a function of the orientation of three HWPs.

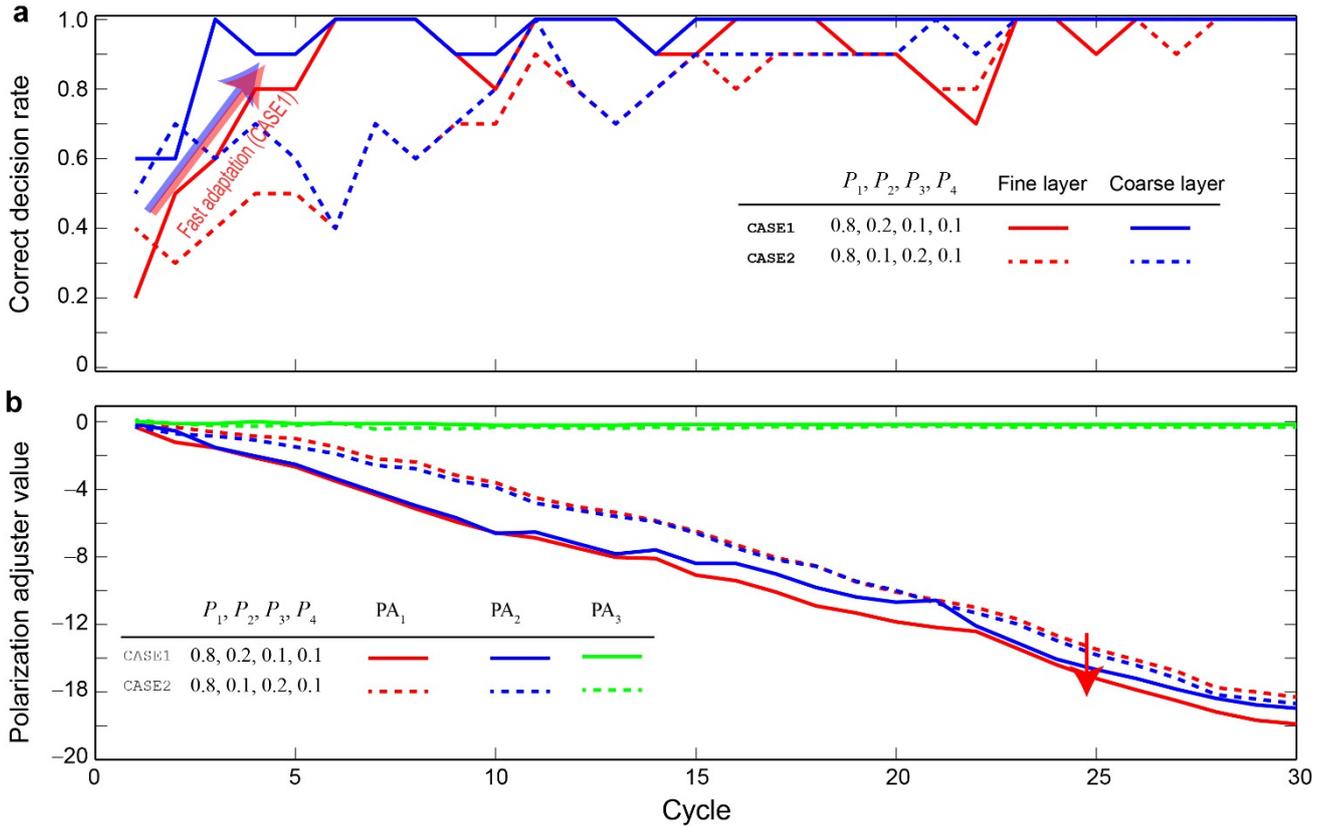

**Figure 3 | Demonstration adaptation at coarse and fine layers of the hierarchical system. (a)** The correct decision rate, based on 30 consecutive plays of slot machines with regard to two problem instances given by CASE 1 and CASE 2, increased over time, demonstrating accurate adaptation of the hierarchical single-photon-based decision maker. The correct decision at the fine layer was to select **slot machine 1,** whereas that at the coarse scale was to select either **slot machine 1 or 2**. Since the coarse scale decision was easier for CASE 1, quicker adaptation was observed in CASE 1. **(b)** Evolution of polarization adjuster values ($PA_i$ ($i=1,\cdots,3$)). $PA_1$ and $PA_2$ for CASE 1 problem decreased rapidly, which was the foundation of the rapid adaptation observed in (a). $PA_3$ stayed about zero for both cases, since the reward probability differences between $P_3$ and $P_4$ were zero or very small, as well as the fact that **slot machines 3** and **4** were not chosen as time elapsed.

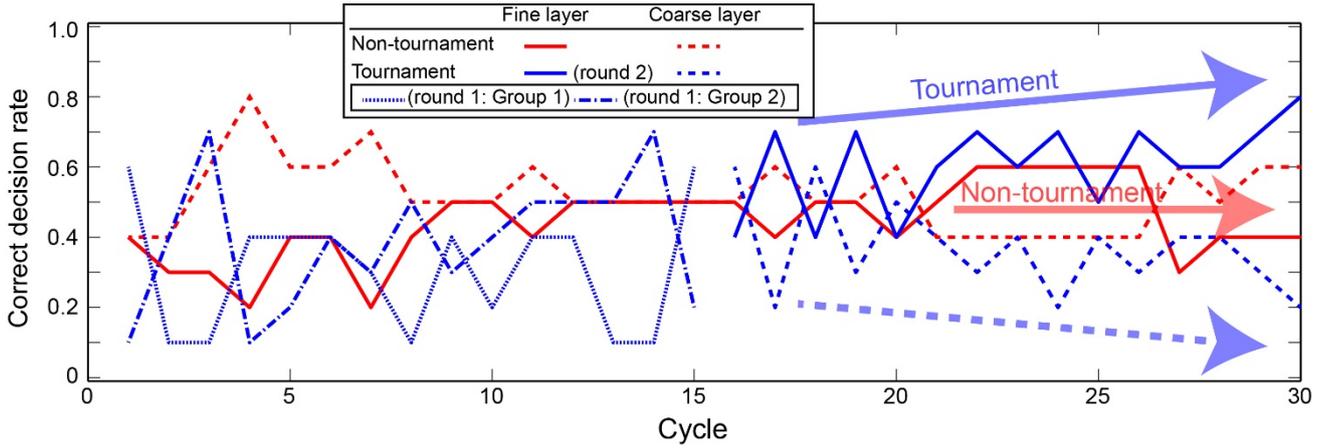

**Figure 4 | Demonstration of the *tournament* method to solve *contradictory* problems at coarse and fine scales.** The optimal solution, or the highest-reward probability slot machine, may *not* belong to the higher-reward probability group at the coarse scale, referred to as contradictory problems, such as $\{P_1, P_2, P_3, P_4\} = \{0.7, 0.5, 0.9, 0.1\}$ (CASE 3). The best option was slot machine 3 ($P_3 = 0.9$), but $P_1 + P_2 > P_3 + P_4$ means that Group 1 (**slot machines 1** and **2**) were better at the coarse scale. The tournament method derived the global optimal, whereby the fine scale local maximum was selected in the first round, followed by the second round, where the global maximum was derived by comparing the winners of the first round. The blue lines show the correct selection rate by the tournament method, which increased over time in the second round, whereas the non-tournament method, depicted by red lines, had difficulty in finding the solution.

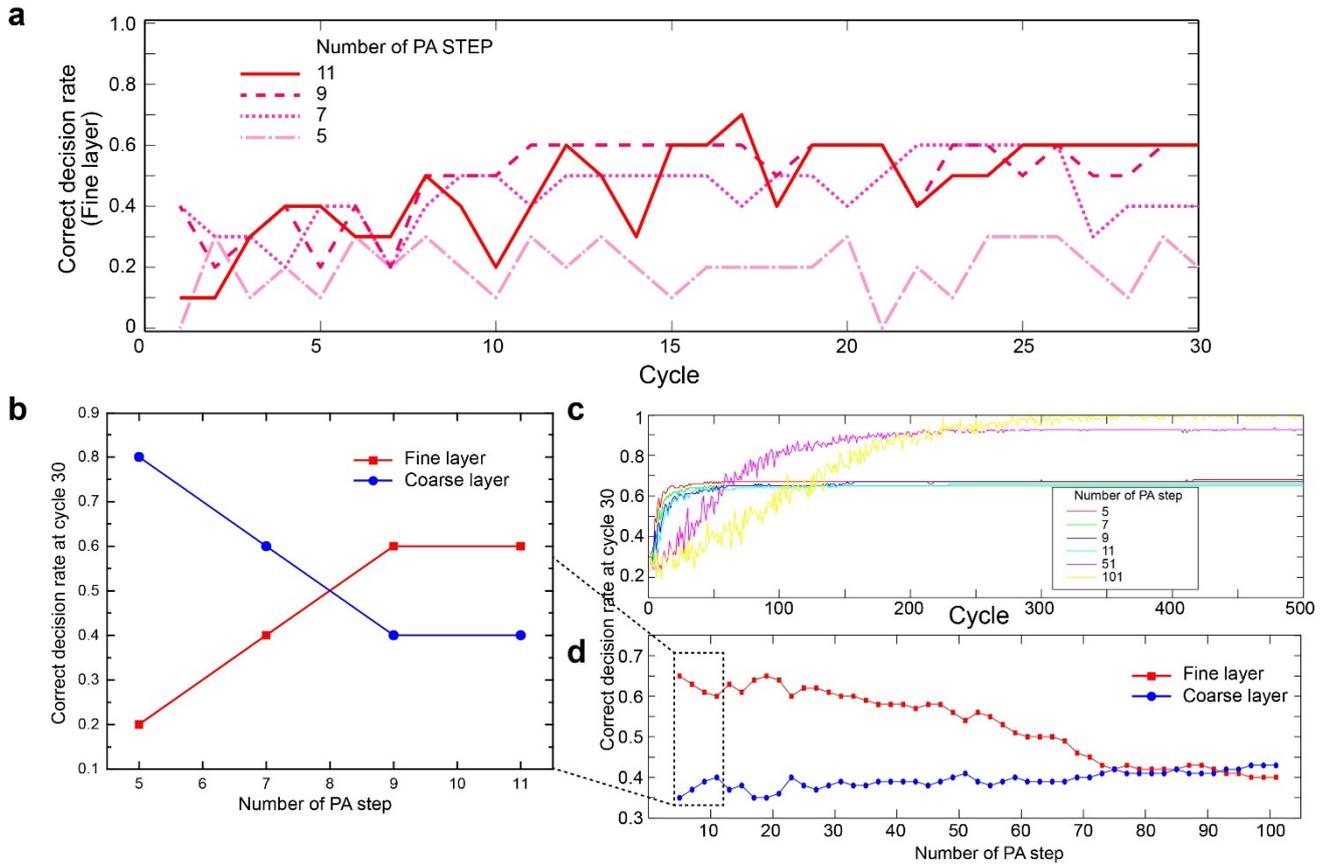

**Figure 5 | Adequate exploitation of probabilistic attribute of single photons allowed direct solution of contradictory problems**. **(a, b)** By enhancing the resolutions of the PAs or the number of steps in the PAs, the probabilistic nature of the single photon was enhanced. By increasing the number of PA steps from five to 11, the correct selection rate of a contradictory problem increased *without* employing the tournament methods. **(c)** Numerical simulation of the correct decision rate revealed that the collect decision rate approached unity by increasing the PA resolution at the expense of slow adaptation. **(d)** The correct decision rate at cycle 30 as a function of the number of PA steps. The trend of PA resolution agreed well with the experiment (b) and the simulation (d).

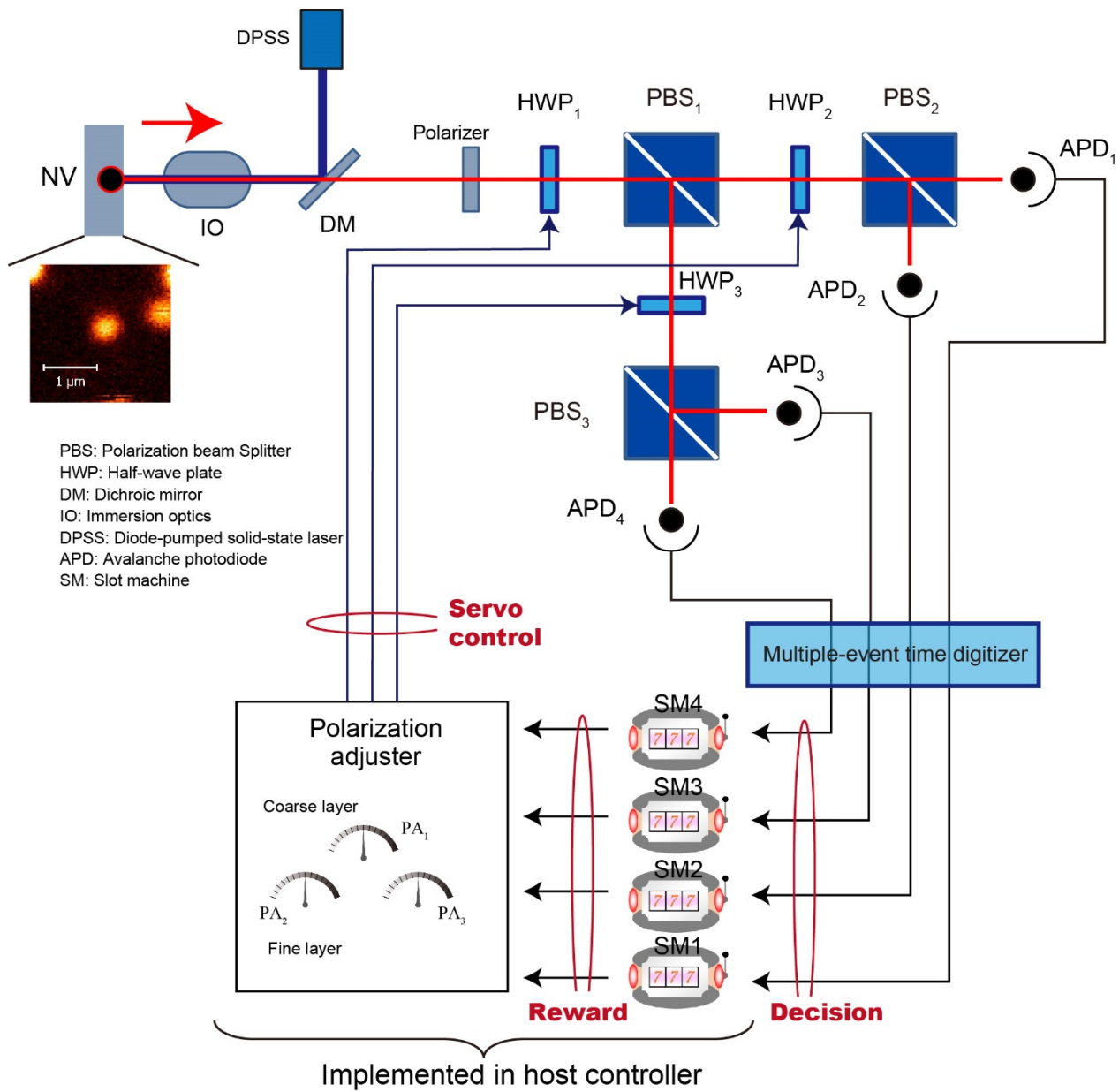

**Supplementary Figure 1 | Schematic diagram of the experimental setup for the hierarchical architecture for reinforcement learning based on single photons.** A nanodiamond is excited by a diode-pumped solid-state laser (DPSS) via a dichroic mirror (DM) and an immersion optics (IO). Single photons, emitted from the NV centre in a nanodiamond, impinges on a polarizer. They are directed to one of four photodetectors (APD$_i$ ($i = 1, \cdots, 4$)) through three half-wave plates (HWP$_i$ ($i = 1, 2, 3$)) and polarized beam splitters (PBS$_i$ ($i = 1, 2, 3$)) arranged in a hierarchical structure. The photon arrival time is measured through a multiple-event time digitizer. The four slot machines and three polarization adjusters (PA$_i$ ($i = 1, 2, 3$)) are implemented in the host controller. The orientations of the half-wave plates are configured by rotary positioners, so that the single photon was more likely to be detected by the higher-reward-probability slot machine.

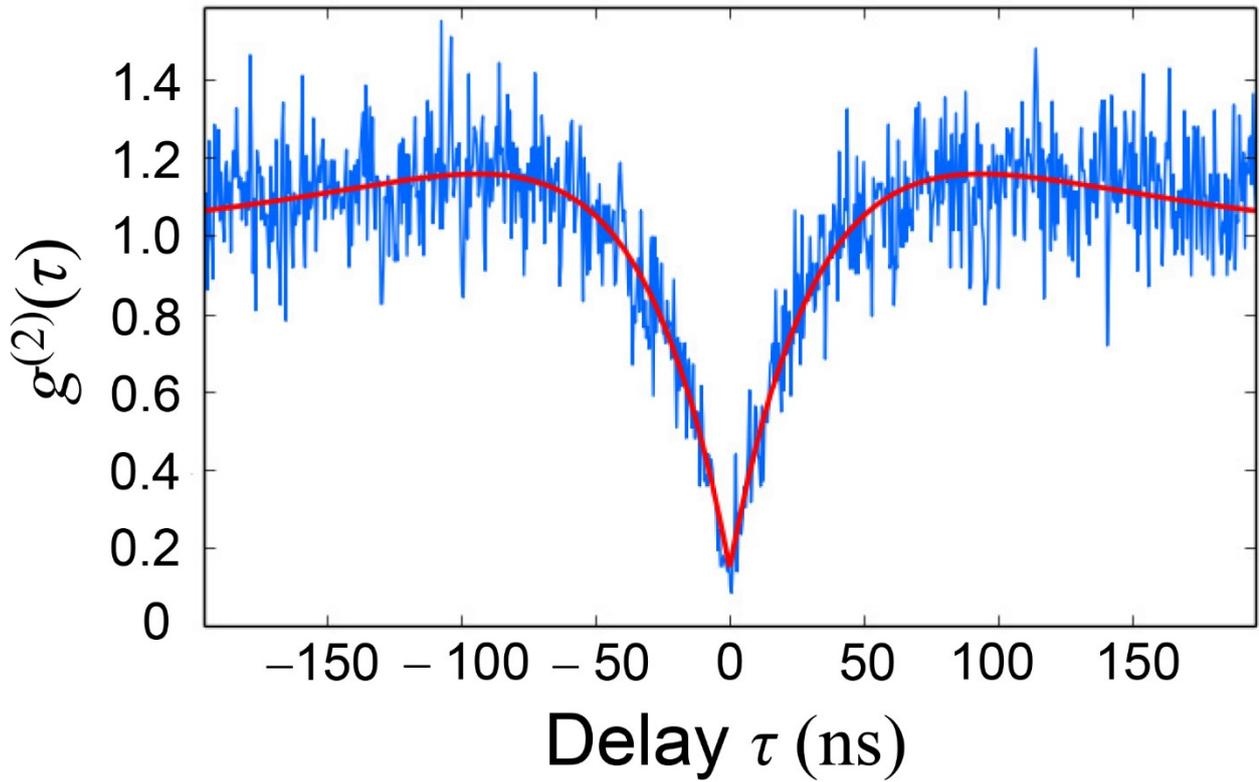

**Supplementary Figure 2 | Second-order photon–intensity correlation measurement of the single NV centre used in the hierarchical reinforcement learning demonstrations.** The red curve is fit to three-level model.